# Efficient Lung Cancer Image Classification and Segmentation Algorithm Based on Improved Swin Transformer

Ruina Sun [1*]   Yuexin Pang[2]

[1] School of Software, North University of China,Taiyuan,030051,China
[2] School of Instrument and Electronics, North University of China,Taiyuan,030051,China

Communication email:1913040602@st.nuc.edu.cn

## Abstract

With the development of computer technology, various models have emerged in artificial intelligence. The transformer model has been applied to the field of computer vision (CV) after its success in natural language processing (NLP). Radiologists continue to face multiple challenges in today's rapidly evolving medical field, such as increased workload and increased diagnostic demands. Although there are some conventional methods for lung cancer detection before, their accuracy still needs to be improved, especially in realistic diagnostic scenarios. This paper creatively proposes a segmentation method based on efficient transformer and applies it to medical image analysis. The algorithm completes the task of lung cancer classification and segmentation by analyzing lung cancer data, and aims to provide efficient technical support for medical staff. In addition, we evaluated and compared the results in various aspects. For the classification mission, the max accuracy of Swin-T by regular training and Swin-B in two resolutions by pre-training can be up to 82.3%. For the segmentation mission, we use pre-training to help the model improve the accuracy of our experiments. The accuracy of the three models reaches over 95%. The experiments demonstrate that the algorithm can be well applied to lung cancer classification and segmentation missions.

## 1. Introduction

In recent years, computer vision has developed rapidly, and it is increasingly being used in a variety of industries. Using computer vision technology for medical detection has also become a trend. Here, the specific application to lung cancer detection is equally valuable for research. The following are the reasons for this.

First, existing studies [1] found that exposure does, such as air pollution and haze index, have a positive relation-ship with the incidence of lung cancer. In addition, Huang et al. [2] found that lung cancer may be increased by long-term exposure to air pollution.

Second, according to a survey [3], the rate of lung cancer misdiagnosis is significant. Clinical medicine has shown that lung cancer symptoms become apparent only in the later stages of the disease when the tumor has progressed widely, making it difficult to diagnose lung cancer at an early stage solely based on a doctor's visual examination of CT images. Besides, physicians are overworked with an average of over 10,000 photographs to read each day. As a result, there is also a greater likelihood of clinician misdiagnosis.

Computer vision technology has made significant advances in medical image analysis thanks to the success of deep learning, and it can now handle a wide range of automated image analysis tasks. One of the most notable applica-tions is the diagnosis of lung cancer using computed tomography imaging.

For research in image detection, we will first discuss CNNs. Convolutional neural networks (CNNs) are artificial neural networks specifically designed to process images [4] and have been influential in image detection research. Alex et al. [5] developed a "large-scale, deep convolutional neural network" (AlexNet). In theory, CNNs allow for the classifi-cation of individual pixels in an image, but CNN training is time-consuming and expensive. Although CNNs offer au-tomatic cost savings compared to traditional techniques, the ensemble layer reduces the image resolution, while the ful-ly connected layer limits the input size to a fixed number of nodes. The attention mechanism is built to address the drawbacks of CNNs. To save computational effort, we introduce the localization of the CNN convolution process and confine attention computation to windows. Google proposed the transformer model in 2017 [6]. They replaced the cir-cular structure in the original Seq2Seq model with attention, which gave a huge shock to the field of natural language processing (NLP). As research evolved, transformers and other related techniques spread from NLP to other fields, such as computer vision (CV). Transformer-based models have shown competitive or even better performance on various vi-sion benchmarks compared to other network types, such as convolutional and recursive networks. Lung cancer identi-fication, which is unique to the medical imaging field, performs equally well in this area.

Applications of computer vision in medical image analysis include classification and segmentation missions, and research in this area has also progressed. In this study, we provide an in-depth study of these two elements, after which we provide a method for detecting lung cancer using ViT for image classification and segmentation.

The purpose of the classification mission is usually to classify the case images (e.g., lung cancer image classifica-tion). Classification mission as an auxiliary diagnosis means can provide some effective suggestions for doctors, which can assist the doctor to improve the speed and accuracy of clinic diagnosis. Kingsley Kuan et al. [7] developed a nodule classifier which is concluded the framework for computer-aided lung cancer diagnosis. The nodules' classifier analyzes the detector output and determines if the nodule is malignant or benign. Pouria Moradi et al. [8] proposed a 3D Convo-lutional Neural Network, whose main aim is to enhance the accuracy of classification.Wei Shen et al. [9] proposed a hierarchical learning framework—Multi-scale Convolutional Neural Networks (MCNN), which can extract discriminative characteristics from alternatingly stacked layers to capture nodule heteroge-neity. Their method can classify malignant and benign nodules effectively absent segmentation of nodules. Using re-stricted chest CT data, Yutong Xie et al. [10] proposed a multi-view knowledge-based collaborative (MV-KBC) deep model to distinguish malignant from benign nodules. The MV-KBC model can reach excellent accuracy, according to their findings.

Because lung cancer does not show symptoms until it has spread, detecting and accurately diagnosing possibly malignant lung nodules early in their formation would improve treatment efficacy and thereby minimize lung cancer mortality. In lung cancer detection, we can accurately infer the location of the nodules by the segmentation mission. Several researchers regard [11-14] the segmentation mission as a classification mission of voxel by voxel. MV-CNN, proposed by Shuo Wang et al. [15], is a CNN-based architecture for lung nodule segmentation [27]. They extract three multi-scale patches from sagittal, coronal, and axial centered on this voxel as input to the CNN model and predict if the voxel corresponds to the nodule when given a voxel in a CT scan.

Vision Transformer (ViT) [16] is the first paper to show how transformers can 'completely' replace traditional con-volutions in deep neural networks on large-scale computer vision datasets. Pre-training on a large proprietary dataset of photos gathered by Google and afterward fine-tuned to downstream identification benchmarks is critical, as pre-training on a medium-range dataset would not produce state-of-the-art results with a ViT.

However, vision trans-former has not been applied in the classification and segmentation of lung cancer, so we come up with this research. As with deeper layers of ViTs, the self-attention mechanism cannot learn effective concepts for feature representation, pre-venting the model from achieving the desired performance boost. Daquan Zhou et al. [17] developed Re-attention, a simple yet effective method for re-generating attention maps to boost their diversity at different layers with little compu-tation and memory cost. Ze Liu [19] et al. presented Swin Transformer, a new vision transformer that can serve as a general-purpose backbone for computer vision. Swin Transformer, which they proposed, includes a shift win-dows-based hierarchical transformer, allowing it to be used for a wide range of vision missions. In addition, vision transformer makes significant advances in medical picture segmentation. Hu Cao et al. [18] introduced Swin-Unet, an Unet-like pure Transformer for medical picture segmentation. To extract context characteristics, they use a hierarchical Swin Transformer with shifted windows as the encoder, and to restore spatial resolution, they adopt a symmetric Swin Transformer-based decoder with a patch expanding layer that they devised.

Given Swin Transformer has potential in the medical image field, we will further try to use it in lung cancer recog-nition in this paper.

In this study, we employ the layered design of the Swin Transformer method with the sliding window operation proposed by Ze Liu et al. [19] for lung cancer detection, a method that has shown revolutionary performance gains over previous methods in the computer vision (CV) field. The specific work we did in the experiment comprised two main parts: the classification mission and the segmentation mission.

In the classification mission, we enlarge the feature part of the original dataset and re-cut the image to get the new dataset. There are two classes in the new dataset — lung nodules and non-lung nodules. Then we train on the Swin Transformer network using two training settings—regular training and pre-training. We get the max accuracy can reach 82.3% for regular training of the Swin-T model, and pre-training of the Swin-B models. In the segmentation mission, the Swin Transformer network has better performance. We slice the images and labels in three directions (x-direction, y-direction, and z-direction). Then we automatically filter the labels with nodules by programming and match them to the images. The results are impressive: the accuracy of the three models can reach over 95%.

## 2. Materials and Methods

### 2.1 Framework Description

To address recognizing early lung cancer, we conduct some lung cancer research using Swin Transformer. Figure 1 depicts the Swin Transformer framework, which comprises three phases: image processing, attention blocking, and downstream operations.

CNN for image processing is to directly treat the image as a matrix for convolution operation, however, the trans-former is originally from NLP and is used to process natural language sequences. It is difficult to be used directly for image feature extraction as CNN. Therefore, we adopt patching operations, which include patch embedding, patch merging, and mask.

#### 2.1.1 Patch Embedding

The role of patch partition is to convert RGB maps into non-overlapping patch blocks. The size of the patch here is 4*4, multiplied by the corresponding RGB channels to get a size of 4*4*3=48.

A feature matrix is obtained by projecting the processed patches to the specified dimensions.

### 2.1.2 Patch Merging

The feature matrix got in the previous step is divided into 2*2 size windows, and the corresponding position of each window is merged, and then the four feature matrices are concatenated after merging.

### 2.1.3 Mask

The mask is constructed so that the window will only do self-attention on the continuous part after the SW-MSA is moved later. The mask slices are shown in Figure 1 (a). The original window is located in the top left matrix, and by a shift to the bottom right.
The formula for the relationship between shift size and window size is as follows.

$$s = \left\lfloor \frac{w}{2} \right\rfloor \quad (1)$$

where *s* is shift size, *w* is the window size.

At this time the area below and to the right can be seen in the window and is not adjacent to the part in the original matrix, so it needs to be divided out with the mask matrix. The vertical slicing area is [0, -window_size), [-window_size, -shift_size), [-shift_size, :), and the horizontal area slicing is the same. For the labeled mask matrix according to the window partition (function window_partition), the idea of the partition is to divide the window size into blocks of $\left\lfloor \frac{H}{w} \right\rfloor$ rows $\left\lfloor \frac{H}{w} \right\rfloor$ columns equally and merge the dimension representing the number and the dimension of the batch size. The purpose of this division is to allow the original matrix mask to become partitioned into small windows and counted in window units. After mask slicing as shown in Figure 2 (b).

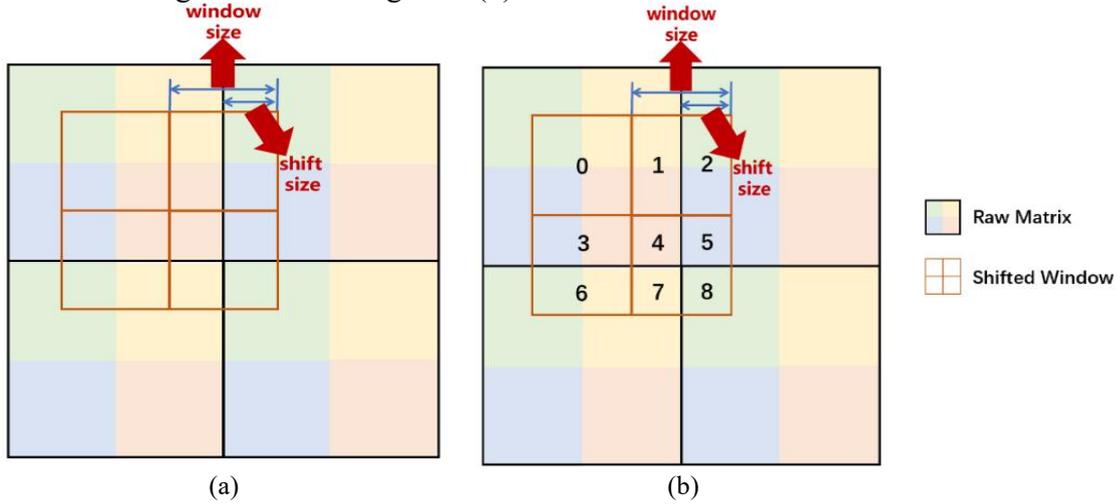

(a)          (b)

Figure 1: Mask slice and after mask slicing. (a) Mask slice; (b) After mask slicing.

## 2.2 Algorithm Design

### 2.2.1 First and Second Stage

In the first stage, we completed the image processing. Like most transformer structures, the RGB lung CT scan im-age is first segmented into a series of non-overlapping patches. In the Swin Transformer [19] setup, each patch has a size of 4*4, and since each pixel has RGB three channel values, each patch has a dimension of 4*4*3 and is finally trans-formed into a C dimensional feature matrix by a linear embedding layer.

The second stage is the Swin Transformer block. Similar to most CNN architectures, Swin Transformer [19] also captures deep characteristics by stacking several blocks. In this paper, we use 4 repeated attention blocks to learn image features. The processed patches are projected to the specified space. We first divide the input feature into C dimension using linear embedding and then send it to Swin Transformer Block. Swin Transformer block comprises a shift win-dow-based MSA and two layers of MLP. Each MSA module and each MLP are preceded by a layer specification (LN) layer, which is followed by a residual connection. After that, the Patch Merging operation first stitches patches in the immediate 2*2 range. This makes the number of patch blocks *H*/8 * *W*/8 and the feature dimension 4C. 4C is compressed into 2C using linear embedding as in stage 1 and then fed into the Swin Transformer block. The combination of these blocks yields a layered representation with the same feature mapping resolution as a normal convolutional network.

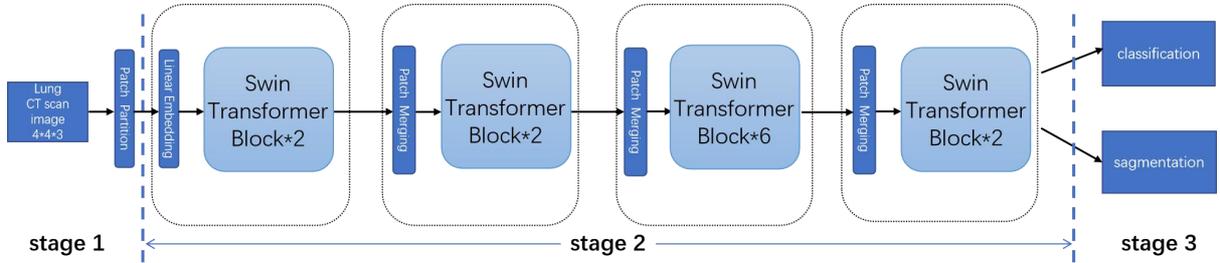

Figure 2: Lung cancer detection process (Swin-T).

## 2.2.2 Self-Attention in Non-Overlapped Windows

Global computation is not ideal for many vision applications that demand huge sets of tokens for dense prediction or representation of high-resolution images since it has quadratic complexity in terms of the number of tokens. Instead, computing self-attention within a local window allows efficient modeling. The photos are segmented consistently and non-overlapping in these windows. The computational complexity of the global MSA module and the windows based on $h * w$ patches pictures is assuming that each window includes $M * M$ patches.

$$\Omega(MSA) = 4hwC^2 + 2(hw)^2 C \tag{2}$$

$$\Omega(W-MSA) = 4hwC^2 + 2M^2 hwC \tag{3}$$

where the former is squared with the number of patches $hw$ and when $M$ is constant, the latter follows a linear path (set to 7 by default). Window-based self-attentiveness is scalable, whereas global self-attentive computing is typi-cally pricey for the huge hardware.

## 2.2.3 Shifted Window Partitioning in Successive Blocks

Since the absence of information exchange between non-overlapping windows undoubtedly limits their modeling capabilities, cross-window connections are introduced. In two successive Swin Transformer blocks, this technique al-ternates W-MSA with SW-MSA. the shifted window division method makes the connection between adjacent non-overlapping windows in the upper layer and increases the perceptual field of view.

With the shifted structure, the Swin Transformer blocks are calculated as follows.

$$\hat{z}^l = W-MSA(LN(z^{l-1})) + z^{l-1},$$

$$z^l = MLP(LN(\hat{z}^l)) + \hat{z}^l,$$

$$\hat{z}^{l+1} = SW-MSA(LN(z^l)) + z^l,$$

$$z^{l+1} = MLP(LN(\hat{z}^{l+1})) + \hat{z}^{l+1}, \qquad (4)$$

where $\hat{z}^l$ and $z^l$ denote the output features of the (S)W-MSA module and the MLP module for block , respec-tively; W-MSA and SW-MSA are window-based multi-head self-attention algorithms that use normal and shifted win-dow partitioning configurations, respectively.

### 2.2.4 Multihead Self-Attention

Multi-head attention mechanism for migration from Transformer to vision. The specific formula is as follows.

$$Attention(Q,K,V) = soft\max\left(\frac{QK^T}{\sqrt{d}} + B\right)V \quad (Q,K,V \in R^{M^2*d}) \qquad (5)$$

where $B$ is the relative position parameter, which is introduced similar to the position embedding in Transformer. d is the size of the dimension corresponding to each head, which serves to balance the size of $QK^T$ and $B$. $Q, K, V$ calcula-tion: for the incoming window information, the corresponding query, key, and value values are obtained after a linear layer.

### 2.2.5 Third Stage

The above introduces how to use Swin Transformer to extract features, and finally, we used Swin Transformer to do the task of classification and segmentation respectively.

The third stage contains two downstream tasks: classification and segmentation. For the classification task, the output dimension is specified as the number of classifications (In our experiments, there were two categories, with and without nodules), and then the output is passed through softmax to obtain the final classification probability. For the segmentation mission, semantic segmentation is made here as the backbone for extracting image features.

### 2.3 Loss Function

The loss function for the classification mission:

$$L = \frac{1}{N}\sum_i L_i = -\frac{1}{N}\sum_i \sum_{c=1}^M y_{ic}\log(p_{ic}) \qquad (6)$$

where $M$ is the number of categories, $y_{ic}$ is the symbolic function (0 or 1), if the true category of the sample $i$ is equal to 2 take 1 otherwise take 0, and $p_{ij}$ is the predicted probability that the observed sample $i$ belongs to category 2.

For the segmentation mission, its loss function is Equation (6) classified by pixel.

# 3. Results

## 3.1 Datasets

### 3.1.1 Classification

The LUNA16 dataset is a public dataset marked by four experienced thoracic radiologists on nodules, which is a part of the LIDC/IDRI [20]. The dataset is composed of 888 low-dose lung CT images, each containing a series of axial sections of the thorax. It is worth mentioning that LUNA16 has selected over three annotations that radiologists agree on and removed tumors less than 3mm, so the dataset does not contain tumors less than 3mm, which is convenient for training, and tumors less than 3mm are difficult for even professional doctors to identify. The number of slices per im-age varies with the machine, layer thickness, and patient. The original image is a 3D image. Each image contains a se-ries of axial sections of the thorax. This three-dimensional image comprises a different number of two-dimensional im-ages.

For preparing benign and malignant classification data of pulmonary nodules:

For each nodule, it is labeled on the 3D image. Our task is how to get the part with nodules from the 3D image and divide it into 2D images. First, we get the center coordinates of all nodes in the annotation file, then we take the (48,48,48) dimensional area image with these coordinates as the 3D image of the candidate lung nodules, and then we slice them one at a time along with the x-direction, y-direction, and z-direction to get our dataset. There are 1351 images of lung nodules and 549714 images of non-lung nodules now. We treat images with nodules as positive and images without nodules as negative. The difference between positive and negative samples is huge. Next, we performed data augmenta-tion. For 1351 lung nodule pictures, we performed a 40-fold expansion (rotation, panning, flipping, etc.). There was also a 20% random selection of 549,714 lung nodule photographs. To make the dataset suitable for Swin Transformer, we batch load all the original 3D datasets and convert them to RGB images.

To do better experiments, we divided the data into three categories: training, testing, and validation. In the training set, the ratio of positive to negative samples is 6:1, in the test set, the ratio of positive to negative samples is 1:1, and in the validation set, the ratio of positive to negative samples is 1.2:1. As seen in Table 1, we get 20565 images of the training dataset, 2571 images of the validated dataset, and 7076 images of the test dataset.

### 3.1.2 Segmentation

For the segmentation mission, we used the MSD dataset [20]. This dataset contains 60 patients, corresponding to 96 CT, and is manually labeled with contour information. The original dataset has a train set and test set in two parts, so we mark the label according to the signature file in the original dataset. And each part contains two classes: 0 and 1.

The first task is to cut the 3D images of the original dataset from x-direction, y-direction, and z-direction. To solve this problem, we slice the images and labels from three directions (x-direction, y-direction, z-direction). We can see Fig-ure 3.

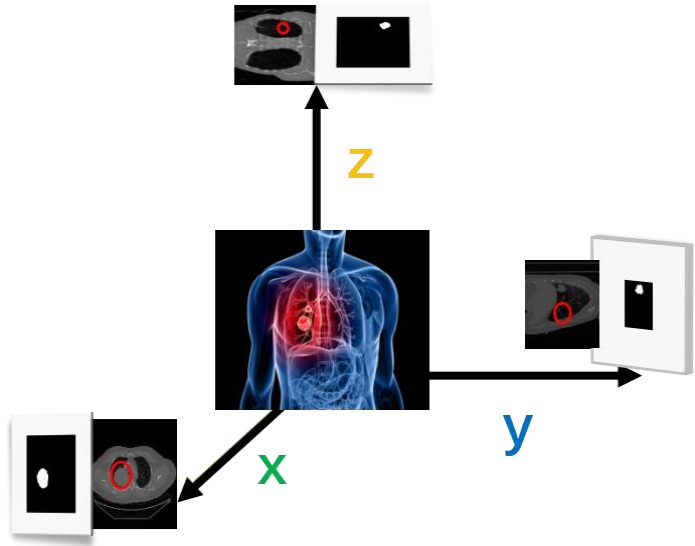

Figure 3: The section of lung nodule from three directions.

We divided the above-processed data into train, test, and validation. As to the specific processing about dividing the dataset, the data set we adopt an 8:1:1 ratio to divide. Finally, as seen in Table 1, we get 22009 images of the training da-taset, 1702 images of the validated dataset, and 1702 images of the test dataset, which can ensure our experiment can get relatively high accuracy.

Table 1: Dataset indication

|  | Classification | Segmentation |
|---|---|---|
| Dataset type | LUNA16 | MSD |
| Train data | 20565 | 22009 |
| Validate data | 2571 | 1702 |
| Test data | 7076 | 1702 |
| Space usage | 18.6 MB | 5.48 GB |

## 3.2 Metric Evaluation

We use top-1 acc and top-5 acc to measure validation set outcomes for medical image classification, where top-1 represents the accuracy of the class with the highest true label prediction probability and top represents the accuracy of the class with the greatest true label prediction probability.

For semantic segmentation of medical images, we use mIoU to evaluate the results, which is a common evaluation method. IoU is calculated for all categories, and then the mean of each category is calculated to get a global evaluation. The specific calculation formula of mIoU is as:

$$MIoU = \frac{1}{k+1} \sum_{i=0}^{k} \frac{p_{ii}}{\sum_{j=0}^{k} p_{ij} + \sum_{j=0}^{k} p_{ji} - p_{ii}} \qquad (7)$$

where $i$ denotes the true value, $j$ denotes the predicted value, $p_{ij}$ denotes the prediction of $i$ to $j$, $k$ denotes the number of categories, and $p_{ii}$ is the number of true samples, which in this paper is the number of successful predic-tions with nodal samples.

In addition, we used the built-in evaluation metrics mAcc and aAcc from mmseg [21,22] to measure the accuracy of the training results.

### 3.3 Experiment Result

### 3.3.1 Classification

We benchmark the Swin Transformer on LUNA16, which involves 12.6 MB training images and 1.58 MB validation images from 2 classes (0 or 1). Considering the particularity of the medical CT dataset, we adopt two training settings to compare which include (1) Regular training; (2) fine-tuning with pre-training. The main instruction is as follows.

For regular training. Our normal training setup in LUNA16 is as follows [24]. A cosine decay learning rate scheduler uses an AdamW [25] for 300 epochs and linear warm-up for 20 epochs. We use a 28-batch size, a 0.001 starting learning rate, and a 0.05 weight decay. Although the training includes the majority of the augmentation and regularization as well as EMA [26], it does not improve performance.

Figure 3 compares Swin-T, Swin-S, and Swin-B in two different resolutions (We denoted these two models respec-tively according to their resolution as Swin-B_224 and Swin-B_384) when it comes to the loss and accuracy for the mis-sions of lung nodule classification. As seen in Figure 4 (a), the overall curve shows a downward trend, among which the Swin-T curve fluctuates significantly, Swin-B_244 curve has the lowest fluctuation range and the best stability. As seen in Figure 4 (b), Swin-T has the highest accuracy of training results, while the other models cannot achieve good training results, which is caused by the different fitness of data set sizes for different models and the lack of pre-training. Mean-while, there is a gradient explosion because the loss values of the training results gradually become NAN. This result is caused by not being pre-trained.

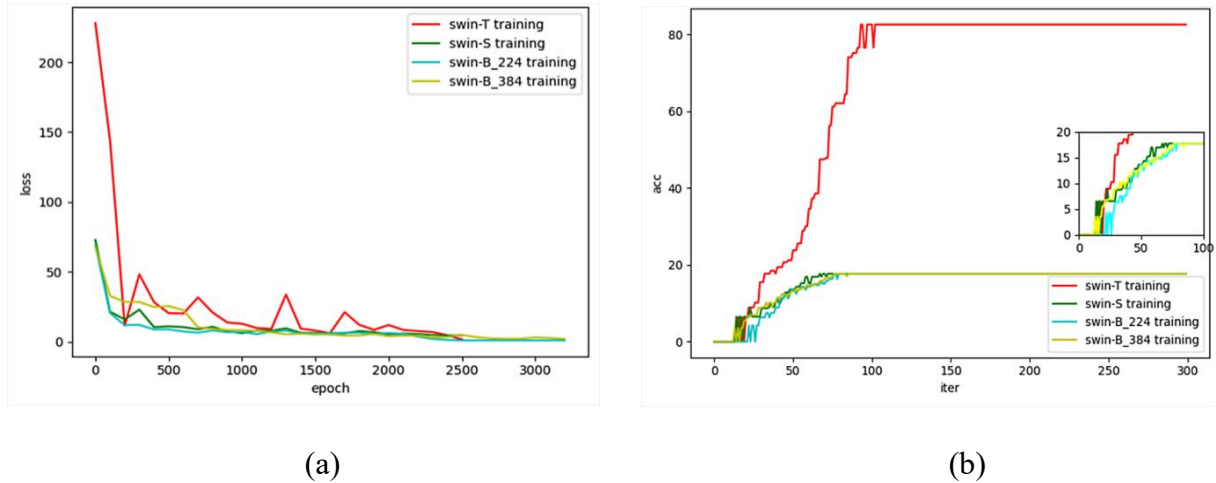

(a)            (b)

Figure 4: Compare loss and accuracy for different models in training.    (a) Loss for training; (b) Accuracy for training.

Table 2 details the result in comparison of different backbones in regular training on LUNA16 classification. The resolutions of the models are shown in Table 2, among which the Swin-B model contains two different resolutions. The results of our validation set are measured by top-1 acc and top-5 acc respectively, where top-1 represents the accuracy of the class with the highest predicted probability for a real label, and top refers to the accuracy of the class with the highest predicted probability for a real label. As seen, Swin-T has the best training effect, with small parameters and high accuracy, which may be caused by the gradient explosion phenomenon and underfitting.

Table 2: Comparison of different backbones in regular training on LUNA16 classification

| Method | Resolution | Top-1 acc | Top-5 acc | Max acc | #params | FLOPs |
|---|---|---|---|---|---|---|
| Swin-T | $224^2$ | 82.26 | 95.9 | 82.3 | 28M | 4.5G |
| Swin-S | $224^2$ | 17.736 | 17.736 | 17.7 | 50M | 8.7G |
| Swin-B | $224^2$ | 17.736 | 17.736 | 17.7 | 88M | 15.4G |
| Swin-B | $384^2$ | 17.736 | 50.0 | 17.7 | 88M | 47.1G |

For pre-training and fine-tuning. Figure 5 compares loss and accuracy for Swin-B in two different resolutions (We denoted these two models respec-tively according to their resolution as Swin-B_224 and Swin-B_384) in pre-training. As seen in Figure 5 (a), the curve showed a downward trend. The initial value of the Swin-B_224 model was high and the decrease range was large, while the loss value of the other model had a relatively small change range. As seen in Figure 5 (b), the accuracy of the two models is depicted in this figure. We can see that the accuracy of the two models finally reaches a good level after pre-training.

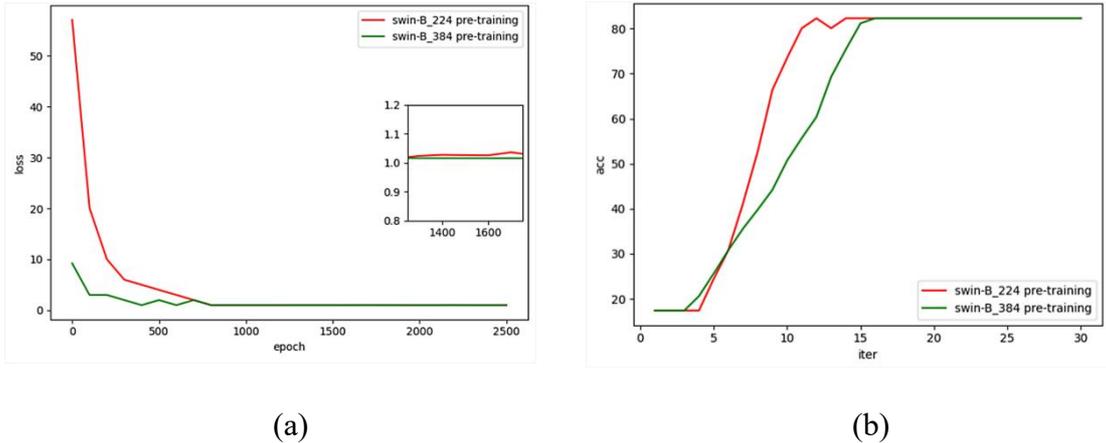

(a)          (b)

Figure 5: Compare loss and accuracy for different models in pre-training. (a) Loss for pre-training; (b) Accuracy for pre-training.

As shown in Table 3, we detail the result in a comparison of different backbones. The table shows that the pre-training model helps improve accuracy to some extent. The two resolutions of Swin-B in Figure 5 (b) refer to specific types of pre-training models [19].

Table 3: Comparison of different backbones in pre-training and fine-tuning on LUNA16 classification.

| Method | Resolution | Top-1 acc | Top-5 acc | Max acc | #params | FLOPs |
|---|---|---|---|---|---|---|
| Swin-B | $224^2$ | 82.264 | 82.264 | 82.3 | 88M | 15.4G |
| Swin-B | $384^2$ | 82.26 | 96.2 | 82.3 | 88M | 47.1G |

To compare the results of the experiment, we pre-train LUNA16 and fine-tuned it. We adopt a linear decay learning rate scheduler with a 5-epoch linear warm-up to run an AdamW optimizer for 30 epochs. We use a 28-batch size, a con-stant learning rate of, $10^{-5}$ and a weight decay of $10^{-8}$.

In regular training, the results are not Unsatisfactory. Only in Swin-T, the max accuracy can reach 82.3%. We can see the result in Table 2: the 1-top accuracy of Swin-S and Swin-B just reach 17.736% and the max accuracy only reach 17.74%.

Differently, the result of pre-training and fine-tuning is an idol. As we expect in Table 3, the test images' max accu-racy can be up to 82.3%, and the gradient explosion phenomenon disappeared.

As shown in Figure 6, shows the comparison of loss curve results of Swin-B_224 and Swin-B_384 models with or without pre-training. The curve decreases rapidly during 0-200, then changes slowly during 200-2000 again, and stabi-lizes after 2000. When the two models are stable, the fluctuation range of the loss curve of the pre-trained model is small.

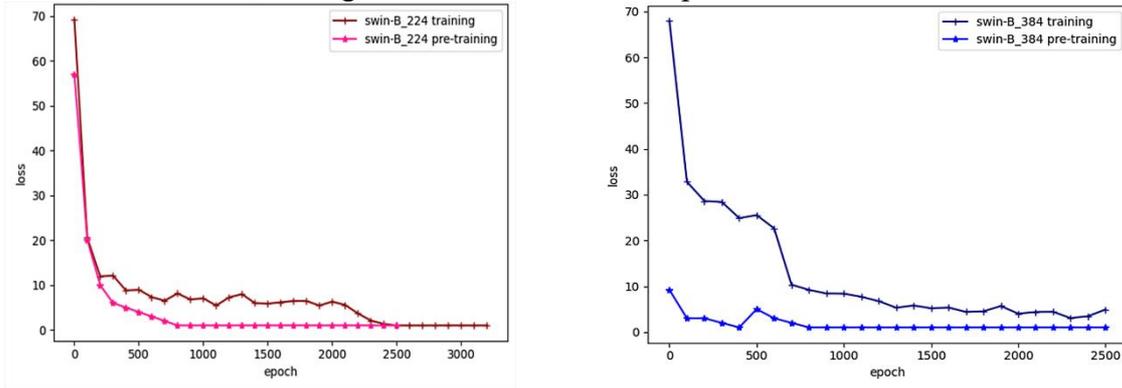

(a)　　　　　　　　　　　　　　　　　　(b)

Figure 6: Compare loss for Swin-B with different image sizes. (a) Compare loss for Swin-B with 224 image size in training and pre-training; (b) Compare loss for Swin-B with 384 image size in training and pre-training.

In addition, we can see the parameter comparison of the four models in Figure 7. As seen in Figure 7 (a), the two resolutions of Swin-B have the same params, but Swin-B_384 shows better performance. As seen in Figure 7 (b), the flops increase with the size of the model.

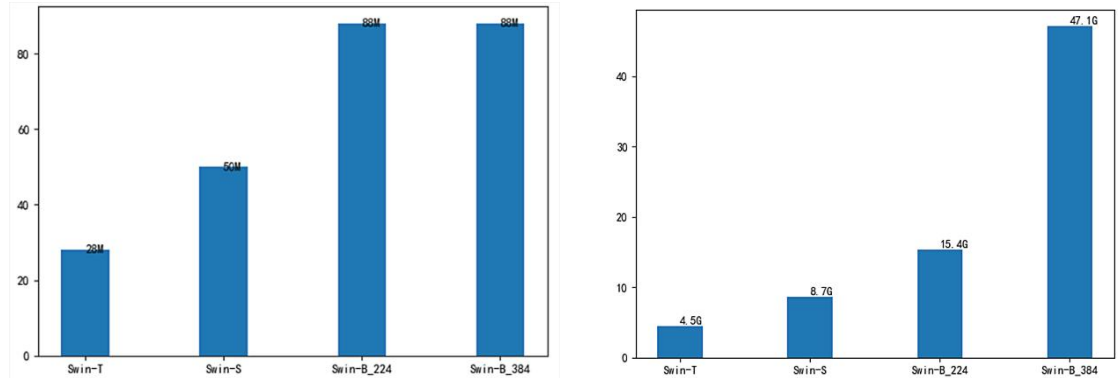

(a)　　　　　　　　　　　　　　　　　　(b)

Figure 7: Comparison of params and flops. (a) Comparison of params; (b) Comparison of flops.

### 3.3.2 Segmentation

For its high efficiency, we employ UperNet [28] in mmseg [22]. The model comes in three sizes: small, small, and tiny. In segmentation trials, we all employ pre-training. We use the AdamW [29] optimizer in training, with a learning rate of  , a weight decay of 0.01, a linear learning rate decay scheduler, and a linear warmup of 1,500 iterations. Models are trained on a single GPU for 40K encounters as a limitation of the experiment. We use the default mmseg set-tings of random horizontal flipping, random rescaling within the ratio range [0.5, 2.0], and random photometric distor-tion for augmentations [19]. For all Swin Transformer models, stochastic depth with a ratio of 0.2 is used.

At the beginning of the segmentation experiments, we meet the loss quickly goes to zero. After the repeated change, the loss still no improvement. Finally, we determined the problem is over-fitting. For solving this sticky problem, we employ data augmentation techniques to generate more similar data from restricted datasets, enrich the distribution of training data, and improve the model's generalizability. We use the data augmentation library to rotate the picture counterclockwise randomly, flip horizontally, flip up and down, and enlarge the image at the same scale. The data augmentation image styles are shown in Figure 8.

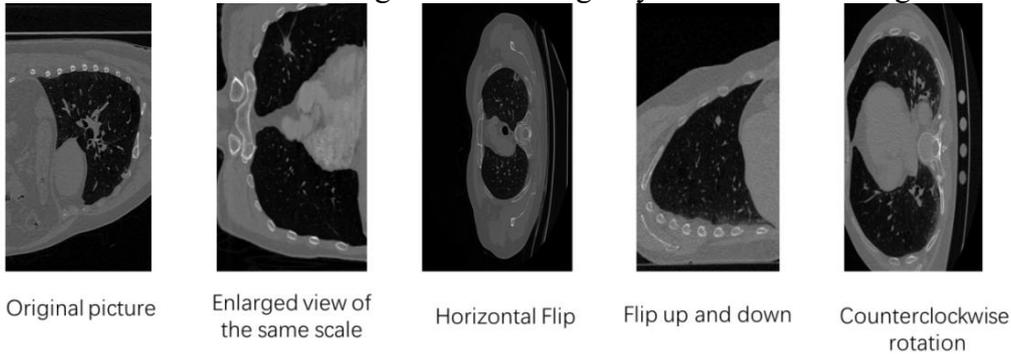

Figure 8: A sample of the data augmentation images

We employ a pre-training strategy to train the dataset to gain the required results because of the big dataset. The pre-training model [19] we used here is the same one used in the classification experiment. We can see the results in Ta-ble 4. The mAcc and aAcc of the three models all reached over 95%. Among them, the Swin-B model has the most out-standing performance, with an accuracy of 99.99%. We use the built-in evaluation metrics mAcc and aAcc from mmseg [22] to measure the accuracy of the training results.

Table 4: Comparison of different backbones in pre-training on MSD segmentation.

| Backbone | Method | #parmas | FLOPs | mIoU | mAcc | aAcc |
|---|---|---|---|---|---|---|
| Swin-T | UPerNet | 60M | 945G | 47.93 | 95.87 | 95.87 |
| Swin-S | UPerNet | 81M | 1038G | 49.94 | 99.87 | 99.87 |
| Swin-B | UPerNet | 121M | 1188G | 49.95 | 99.91 | 99.91 |

The loss curves of the three models are shown in Figure 9. The curves drop rapidly in the process of 0-750, then change slowly in the process of 750-1500, and stabilize after 2000. All three models have small loss values as they con-verge to stability.

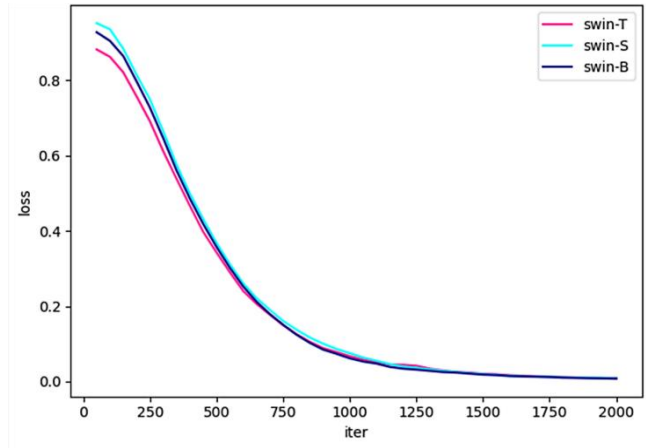

Figure 9: Compare loss for different models

From Figure 10, we can see the comparison of the accuracy of the training results of the three models. Among them, the accuracy of Swin-B is the best, and the accuracy of Swin-T is the worst. The Swin-T curve fluctuates the most and the Swin-B curve is the smoothest.

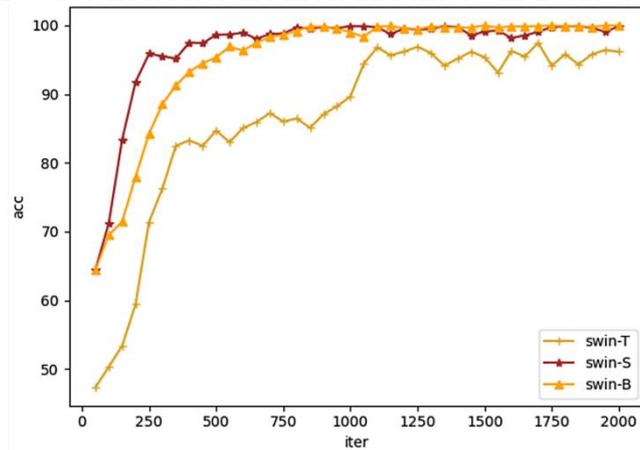

Figure 10: Compare accuracy for different models

As shown in Figure 11, we can see a global evaluation from the mIoU curve, and the Swin-S and Swin-B models re-veal the best effect. The mIoU of Swin-T is the worst.

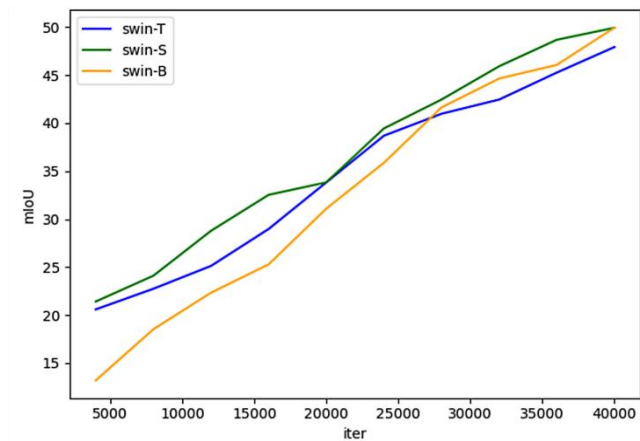

Figure 11: Compare mIoU for different models

In addition, we can get a clear comparison of model parameters params and flops from Figure 12. We can see that the larger the model size is, the larger the values of params and flops are.

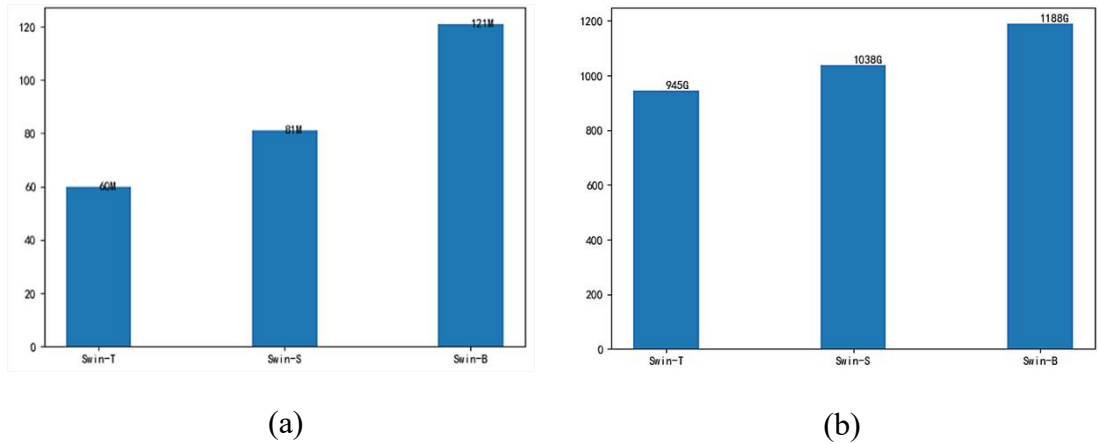

(a)    (b)

Figure 12: Comparison of params and flops. (a) Comparison of params; (b) Comparison of flops.

## 4. Discussion

As it is known to us, transformer has made great progress in traditional image processing technology. In this paper, we innovatively propose a segmentation method based on efficient transformer and applies it to medical image analysis. It's a new visual converter that generates hierarchical feature representations with linear computing complexity about the size of the input image. Compared with other algorithms, our method is more efficient.By completing two missions, Swin Transformer in CT image classification on the LUNA16 dataset and segmentation of the MSD dataset, it is shown that Swin Transformer has desirable results in the specific lung cancer detection domain.

In the future, we will improve Swin Transformer to make it more universally applicable in the field of medical im-aging. In addition, the dataset in this paper was pre-processed and converted into 2D images. Due to the prevalence of 3D medical images, in the future, we will study how Swin Transformer can be applied in 3D medical image classifica-tion and segmentation.

## Conflicts of Interest

The authors declare that there is no conflict of interest regarding the publication of this paper.